\documentclass{llncs}
\usepackage{hyperref}       % hyperlinks
\usepackage{url}            % simple URL typesetting
\usepackage{booktabs}       % professional-quality tables
\usepackage{amsfonts}       % blackboard math symbols
\usepackage{nicefrac}       % compact symbols for 1/2, etc.
\usepackage{microtype}      % microtypography
\usepackage[pdftex]{graphicx}
\usepackage{amsmath}

\begin{document}

\title{Normative Modeling of Neuroimaging Data using Scalable Multi-Task Gaussian Processes}
\titlerunning{Normative Modeling using Multi-Task Gaussian Processes}  % abbreviated title (for running head)
\author{Seyed Mostafa Kia \inst{1,2} \and Andre Marquand \inst{1,2}}
\authorrunning{Kia et al.} % abbreviated author list (for running head)
\tocauthor{Seyed Mostafa Kia, Andre Marquand}
\institute{Department of Cognitive Neuroscience, Radboud University Medical Centre, Nijmegen, The Netherlands \\
%\\ WWW home page: \texttt{http://webpage.html}
\and
Donders Centre for Cognitive Neuroimaging, Donders Institute for Brain, Cognition and Behaviour, Radboud University, Nijmegen, The Netherlands
\email{\{s.kia,a.marquand\}@donders.ru.nl}}

\maketitle              % typeset the title of the contribution
\vspace{-0.5cm}
\begin{abstract}
Normative modeling has recently been proposed as an alternative for the case-control approach in modeling heterogeneity within clinical cohorts. Normative modeling is based on single-output Gaussian process regression that provides coherent estimates of uncertainty required by the method but does not consider spatial covariance structure. Here, we introduce a scalable multi-task Gaussian process regression (S-MTGPR) approach to address this problem. To this end, we exploit a combination of a low-rank approximation of the spatial covariance matrix with algebraic properties of Kronecker product in order to reduce the computational complexity of Gaussian process regression in high-dimensional output spaces. On a public fMRI dataset, we show that S-MTGPR: 1) leads to substantial computational improvements that allow us to estimate normative models for high-dimensional fMRI data whilst accounting for spatial structure in data; 2) by modeling both spatial and across-sample variances, it provides higher sensitivity in novelty detection scenarios.

\vspace{-0.5cm}
\end{abstract}
\section{Introduction} 
\label{sec:introduction}
% Clinical neuroscience and normative modeling
Understanding the underlying biological mechanisms of psychiatric disorders constitutes a significant step toward developing more effective and individualized treatments (\emph{i.e.}, \emph{precision medicine}~\cite{Mirnezami2012preparing}). Recent advances in neuroimaging and machine learning provide an exceptional opportunity to employ brain-derived biological measures for this purpose. While symptoms and biological underpinnings of mental diseases are known to be highly heterogeneous, data-driven approaches play an important role in stratifying clinical groups into more homogeneous subgroups. Currently, off-the-shelf clustering algorithms are the most predominant approaches for stratifying clinical cohorts. However, the high-dimensionality and complexity of data beside the use of heuristics to find optimal clustering solutions negatively affect the reproducibility and reliability of resulting clusters~\cite{marquand2016beyond}. Normative modeling~\cite{marquand2016understanding} offers an alternative approach to model biological variations within clinical cohorts without needing to assume cleanly separable clusters or cohorts. This approach is applicable to most types of neuroimaging data such as structural/functional magnetic resonance imaging (s/fMRI).

% More about Normative modeling and the usage of Gaussian Processes in normative modeling
Normative modeling employs Gaussian process regression (GPR)~\cite{williams1996gaussian} to predict neuroimaging data on the basis of clinical and/or behavioral covariates. GPR, and in general Bayesian inference, can be seen as an indispensable part of the normative modeling as it provides coherent estimates of  predictive confidence. These measures of predictive uncertainty are important for quantifying centiles of variation in a population~\cite{marquand2016understanding}. GPR also provides the possibility to accommodate both linear and nonlinear relationships between clinical covariates and neuroimaging data.

% How multi-task Gaussian processes can help normative modeling? Spatial modelling? Where is the gap?
The variant of GPR originally employed for normative modeling aims to model only a single output variable. Thus in normative modeling, one should independently train separate GPR models for each unit of measurement (\emph{e.g.}, for each voxel in a mass-univariate fashion). Such a simplification ignores the possibility of modeling and capitalizing on the existing spatial structure in the output space. However, GPR can be extended to perform a joint prediction across multiple outputs in order to account for correlations between variables in neuroimaging data (for example different voxels in fMRI data). Boyle and Frean~\cite{boyle2005multiple} proposed to employ convolutional processes to express each output as the convolution between a smoothing kernel and a latent function. This idea is later adopted by Bonilla \emph{et al.}~\cite{bonilla2008multi} to extend the classical single-task GPR (STGPR) to multi-task GPR (MTGPR) by coupling a set of latent functions with a shared GP prior in order to directly induce correlation between output variables (tasks). They proposed to disentangle the full cross-covariance matrix into the Kronecker product of the sample (in input space) and task (in output space) covariance matrices. This technique provides the possibility to model both across-sample and across-task variations. Despite its effectiveness in modeling structures in data, MTGPR comes with extra computational overheads in time and space, especially when dealing with high-dimensional neuroimaging data. We briefly review recent efforts toward alleviating these computational burdens. 
\vspace{-0.25cm}
\subsection{Toward Efficient and Scalable MTGPR} 
\label{subsec:related_work}
% A brief history on the multi--output Gaussian process regression and its high computational complexity 
For $N$ samples and $T$ tasks, the time and space complexity of MTGPR are $\mathcal{O}(N^3 T^3)$ and $\mathcal{O}(N^2 T^2)$, respectively. These high computational demands (compared to STGPR with $\mathcal{O}(N^3 T)$ and $\mathcal{O}(N^2 T)$) are mainly due to the need for computing the inverse cross-covariance matrix in learning and inference phases. In neuroimaging problems that we consider, these can both be relatively high where $N$ in generally in the order of $10^2-10^4$ and $T$ is in the order of $10^4-10^5$ or even higher. Therefore, improving the computational efficiency of MTGPR is crucial for certain problems, and there have been several approaches proposed for this in the machine learning literature~\cite{quinonero2007approximation,alvarez2011computationally}. Here we briefly review two main directions to address the computational tractability issue of MTGPR.

% Some solutions to high complexity
In the first set of approaches, approximation techniques are used to improve estimation efficiency. Bonilla \emph{et al.}~\cite{bonilla2008multi} made one of the earliest efforts in this direction, in which they proposed to use Nystr\"{o}m approximation on $M$ inducing inputs~\cite{quinonero2007approximation} out of $N$ samples in combination with the probabilistic principal component analysis, in order to approximate reduced $M$-rank and $P$-rank sample and task covariance matrices, respectively. Their approximation reduced the time complexity of hyperparameter learning to $\mathcal{O}(N T M^2 P^2)$. Elsewhere, Alvarez and Lawrence~\cite{alvarez2009sparse} proposed to approximate a sparse version of MTGPR, assuming conditional independence between each output variable with all others given the input process. This assumption besides using $M$ out of $N$ input samples as inducing inputs reduces the computational complexity of MTGPR to $\mathcal{O}(N^3 T+ N T M^2)$ and $\mathcal{O}(N^2 T+N T M)$ in time and storage, where for $N=M$ is the same as a set of $T$ independent STGPRs. Alvarez \emph{et al.} in~\cite{alvarez2010efficient} extended their previous work by developing the concept of inducing function rather than inducing input. Their new approach so-called variational inducing kernels achieves time complexity of $\mathcal{O}(N T M^2)$.

The second set of approaches utilize properties of Kronecker product~\cite{loan2000ubiquitous} to reduce the time and space complexity in computing the exact (and not approximated) inverse covariance matrix. Stegle \emph{et al.}~\cite{stegle2011efficient} proposed to use these properties in combination with eigenvalue decomposition of input and task covariance matrices for efficient parameter estimation, and likelihood evaluation/optimization in MTGPR. In this method, the joint covariance matrix is defined as a Kronecker product between the input and task covariance matrices. This approach reduces the time and space complexity of MTGPR to $\mathcal{O}(N^3+T^3)$   and $\mathcal{O}(N^2+T^2)$, respectively. To account also for structured noise, Rakitsch \emph{et al.}~\cite{rakitsch2013all} extended this method by using two separate Kronecker products for the signal and noise. Importantly,  this provides a significant reduction in computational complexity using all samples (\emph{i.e.}, not just inducing inputs), and is exact in the sense that it does not require any approximation or relaxing assumptions.
\vspace{-0.5cm}
\subsubsection{Our contribution:}
In spite of all aforementioned efforts, applications of MTGPR in encoding neuroimaging data from a set of clinically relevant covariates remained very limited, mainly due to the high dimensionality of the output space (\emph{i.e.}, very large $T$). Our main contribution in this text addresses this problem and extends MTGPR to the normative modeling of neuroimaging data. To this end, we use a combination of low-rank approximation of the task covariance matrix with algebraic properties of Kronecker product in order to reduce the computational complexity of MTGPR. Furthermore, on a public fMRI dataset, we show that: 1) our method makes MTGPR possible on very high-dimensional output spaces; 2) it enables us to model both across-space and across-subjects variations, hence provides more sensitivity for the resulting normative model in novelty detection.
\vspace{-0.5cm}
\section{Methods} \label{sec:methods}
\vspace{-0.25cm}
\subsection{Notation} 
\label{subsec:notation}
Boldface capital letters, $\mathbf{A}$, and capital letters, $A$, are used to denote matrices and scalar numbers. We denote the vertical vector which is resulted from collapsing columns of a matrix $\mathbf{A} \in \mathbb{R}^{N \times T}$ with $vec(\mathbf{A}) \in \mathbb{R}^{N T}$. In the remaining text, we use $\otimes$ and $\odot$ to respectively denote Kronecker and the element-wise matrix products. We denote an identity matrix by $\mathbf{I}$; and the determinant, diagonal elements, and the trace of matrix $\mathbf{A}$ with $\left | \mathbf{A} \right |$, $diag(\mathbf{A})$, and $Tr[\mathbf{A}]$, respectively.
\vspace{-0.5cm}
\subsection{Scalable Multi-Task Gaussian Process Regression} 
\label{subsec:SMTGP}
Let $\mathbf{X} \in \mathbb{R}^{N \times F}$ be the input matrix with $N$ samples and $F$ covariates. Let $\mathbf{Y} \in \mathbb{R}^{N \times T}$ represent a matrix of response variables with $N$ samples and $T$ tasks (here, neuroimaging data with $T$ voxels). The multi-task Kronecker Gaussian process model (MT-Kronprod)~\cite{stegle2011efficient} is defined as:
\vspace{-0.15cm}
\small
\begin{eqnarray} \label{eq:MT-kronprod}
p(\mathbf{Y} \mid \mathbf{D},\mathbf{R},\sigma^2) = \mathcal{N}(\mathbf{Y} \mid \mathbf{0}, \mathbf{D} \otimes \mathbf{R} + \sigma^2 \mathbf{I}) \quad ,
\end{eqnarray}
\normalsize
\noindent where $\mathbf{D} \in \mathbb{R}^{T \times T}$ and $\mathbf{R} \in \mathbb{R}^{N \times N}$ are respectively the task and sample covariance matrices (here, modeling correlations across voxels and samples separately). Despite its effectiveness in modeling both samples and tasks variations, the application of MT-Kronprod is limited when dealing with very large output spaces, such as neuroimaging data, mainly due to the high computational complexity of matrix diagonalisation operations in the optimization and inference phases. We propose to address this problem by using a low-rank approximation of $\mathbf{D}$.  

Let $\Phi: \mathbf{Y} \to \mathbf{Z}$ be an orthogonal linear transformation, \emph{e.g.}, principal component analysis (PCA), that transforms $\mathbf{Y}$ to a reduced latent space $\mathbf{Z} \in \mathbb{R}^{N \times P}$, where $P < T$, and we have $\mathbf{Z} = \Phi(\mathbf{Y}) = \mathbf{Y}\mathbf{B}$. Here, columns of $\mathbf{B} \in \mathbb{R}^{T \times P}$ represent a set of $P$ orthogonal basis functions. Assuming a zero-mean matrix normal distribution for $\mathbf{Z}$, by factorizing its rows and columns we have:
\small
\begin{eqnarray} \label{eq:MND_Z}
p(\mathbf{Z} \mid \mathbf{C},\mathbf{R})= \mathcal{MN}(\mathbf{0},\mathbf{C} \otimes \mathbf{R})=\frac{\exp(-\frac{1}{2}Tr[\mathbf{C}^{-1}\mathbf{B}^\top\mathbf{Y}^\top\mathbf{R}^{-1}\mathbf{Y}\mathbf{B}])}{\sqrt{(2\pi)^{NP}\left | \mathbf{C} \right |^P \left | \mathbf{R} \right |^N}} \quad ,
\end{eqnarray} 
\normalsize
\noindent where $\mathbf{C} \in \mathbb{R}^{P \times P}$ and $\mathbf{R} \in \mathbb{R}^{N \times N}$ are column and row covariance matrices of $\mathbf{Z}$. Using the trace invariance property under cyclic permutations, the noise-free multivariate normal distribution of $\mathbf{Y}$ can be approximated from Eq.~\ref{eq:MND_Z}:
\small
\begin{eqnarray} \label{eq:MND_Y}
p(\mathbf{Y} \mid \mathbf{D},\mathbf{R}) \approx p(\mathbf{Y} \mid \mathbf{C},\mathbf{B},\mathbf{R}) = \frac{\exp(-\frac{1}{2}Tr[\mathbf{B}\mathbf{C}^{-1}\mathbf{B}^\top \mathbf{Y}^\top\mathbf{R}^{-1}\mathbf{Y}])}{\sqrt{(2\pi)^{NT}\left | \mathbf{BCB}^\top \right |^T \left | \mathbf{R} \right |^N}} \quad ,
\end{eqnarray}
\normalsize
\noindent where $\mathbf{D}$ is approximated by $\mathbf{B}\mathbf{C}\mathbf{B}^\top$. Our scalable multi-task Gaussian process regression (S-MTGPR) model is then derived by marginalizing over noisy samples:
\small
\begin{eqnarray} \label{eq:kronecker_GP}
p(\mathbf{Y} \mid \mathbf{D},\mathbf{R}, \sigma^2) \approx p(\mathbf{Y} \mid \mathbf{C},\mathbf{B}, \mathbf{R}, \sigma^2) = \mathcal{N}(\mathbf{Y} \mid \mathbf{0}, \mathbf{BCB}^\top \otimes \mathbf{R} + \sigma^2 \mathbf{I}) \quad .
\end{eqnarray}
\normalsize
\vspace{-1cm}
\subsubsection{Predictive Distribution:} 
\label{subsubsec:prediction}
Following the standard GPR framework~\cite{williams1996gaussian} and setting $\tilde{\mathbf{D}}=\mathbf{BCB}^\top$, the mean and variance of the predictive distribution of unseen samples, \emph{i.e.}, $p(vec(\mathbf{Y})^* \mid vec(\mathbf{M^*}), \mathbf{V}^*)$, can be computed as follows:
\small
\begin{subequations} \label{eq:predictive_distribution}
\begin{align}
& vec(\mathbf{M^*}) = (\tilde{\mathbf{D}} \otimes \mathbf{R}^*)(\tilde{\mathbf{D}} \otimes \mathbf{R} + \sigma^2 \mathbf{I})^{-1} vec(\mathbf{Y}), \\
& \mathbf{V}^* = (\tilde{\mathbf{D}} \otimes \mathbf{R}^{**})-(\tilde{\mathbf{D}} \otimes \mathbf{R}^*)(\tilde{\mathbf{D}} \otimes \mathbf{R} + \sigma^2 \mathbf{I})^{-1}(\tilde{\mathbf{D}} \otimes \mathbf{R}^{* \top}),
\end{align}
\end{subequations}
\normalsize
\noindent where $\mathbf{R}^{**} \in \mathbb{R}^{N^* \times N^*}$ is the covariance matrix of $N^*$ test samples , and $\mathbf{R}^* \in \mathbb{R}^{N^* \times N}$ is the cross-covariance matrix between test and training samples.   
\subsubsection{Efficient Prediction and Optimization:} 
\label{subsubsec:optimization}
For efficient prediction, and fast optimization of the log-likelihood, we extend the approach proposed in~\cite{stegle2011efficient,rakitsch2013all} by exploiting properties of Kronecker product, and eigenvalue decomposition for diagonalizing the covariance matrices. Then the predictive mean and variance can be efficiently computed by:
\small
\begin{subequations} \label{eq:efficient_prediction}
\begin{align}
\mathbf{M}^* &= \mathbf{R}^* \mathbf{U_R} \mathbf{\tilde Y} \mathbf{U^\top_C} \mathbf{C}\mathbf{B}^\top, \\
\mathbf{V}^* &= (\tilde{\mathbf{D}} \otimes \mathbf{R}^{**})-(\mathbf{BCU_C} \otimes \mathbf{R}^* \mathbf{U_R})\tilde{\mathbf{K}}^{-1}(\mathbf{U_C^\top CB^\top} \otimes \mathbf{U_R^\top R}^{* \top}),
\end{align}
\end{subequations}
\normalsize
\noindent where $\mathbf{C=U_CS_CU_C^\top}$ and $\mathbf{R=U_RS_RU_R^\top}$ are eigenvalue decomposition of covariance matrices, $\tilde{\mathbf{K}} = \mathbf{S_C} \otimes \mathbf{S_R} + \sigma^2 \mathbf{I}$, and $vec(\mathbf{\tilde Y})=diag(\tilde{\mathbf{K}}^{-1}) \odot vec(\mathbf{U_R^\top Y B U_C})$.\footnote{\scriptsize See supplementary materials for more descriptive derivations of all equations.\normalsize} Based on our assumption on the orthogonality of components in $\mathbf{B}$, we set $\mathbf{B}^{-1}=\mathbf{B}^\top$ and $\mathbf{B}^\top \mathbf{B}=\mathbf{I}$. Note that in the new parsimonious formulation, heavy time and space complexities of computing the inverse kernel matrix is reduced to computing the inverse of a diagonal matrix, \emph{i.e.}, reciprocals of diagonal elements of $\tilde{\mathbf{K}}$. For the predictive variance, explicit computation of the Kronecker product is still necessary but this can easily be overcome by computing the predictions in mini-batches. For the negative log marginal likelihood of Eq.~\ref{eq:kronecker_GP}, we have: \vspace{-0.15cm}
\small
\begin{eqnarray} \label{eq:LML}
\mathcal{L}=-\frac{N \times T}{2} \ln(2\pi)-\frac{1}{2}\ln\left | \tilde{\mathbf{K}} \right | - \frac{1}{2} vec(\mathbf{U_R^\top YBU_C})^\top vec(\mathbf{\tilde Y}) \quad .
\end{eqnarray}
\normalsize

The proposed S-MTGPR model has three sets of parameters plus one hyperparameter: 1) reduced task covariance matrix parameters $\Theta_{\mathbf{C}}$, 2) input covariance matrix parameters $\Theta_{\mathbf{R}}$, 3) noise variance $\sigma^2$ that is parametrized on $\Theta_{\sigma^2}$, and 4) $P$ that decides the number of components in $\mathbf{B}$. While the latter should be decided by means of model selection, the first three sets are optimized by maximizing $\mathcal{L}$. 
\vspace{-0.4cm}
\subsubsection{Computational Complexity:} 
\label{subsubsec:complexity}
The time complexity of the proposed method is $\mathcal{O}(N^2 T + N T^2 + N^3 + P^3)$. The first two terms are related to the matrix multiplication in computing the squared term in Eq.\ref{eq:LML}. The last two terms belong to the eigenvalue decomposition of $\mathbf{R}$ and $\mathbf{C}$. The $P^3$ term can be excluded because always $P \leq min(N,T)$. Thus, for $N>T$ and $N<T$ the time complexity is reduced to $\mathcal{O}(N^3)$ and $\mathcal{O}(N T^2)$, respectively. Thus when $N>T$ or $N<T<N^2$, our approach is analytically even faster than the baseline STGPR approach applied independently to each output variable in a mass-univariate fashion. For $N \ll T$, our method is faster than other Kronecker based MTGPRs by a factor of $T/N$. Such improvement not only facilitates the application of MTGPR on neuroimaging data but also it provides the possibility of accounting for the existing spatial structures across different brain regions. In comparison to the related work, the proposed method provides a substantial speed improvement, especially when dealing with a large number of tasks. This is while unlike other approximation approaches, we fully use the potential of all available samples.
\vspace{-0.4cm}
\section{Experiments and Results} \label{sec:results}
\vspace{-0.25cm}
\subsection{Experimental Materials and Setup} \label{subsec:materials_setups}
In our experiments, we use a public fMRI dataset collected for reconstructing visual stimuli (black and white letters and symbols) from fMRI data~\cite{miyawaki2008visual}. In this dataset, fMRI responses were measured while $10 \times 10$ checkerboard patch images were presented to subjects according to a blocked design. Checkerboard patches constituted random (1320 trials) and geometrically meaningful patterns (720 trials). We use the preprocessed data available in Nilearn package~\cite{abraham2014machine} wherein the fMRI data are detrended and masked for the occipital lobe (5438 voxels).\footnote{\scriptsize See \url{http://nilearn.github.io/auto_examples/02_decoding/plot_miyawaki_reconstruction.html}. \normalsize} Whilst our approach is quite general, we demonstrate S-MTGPR by simulating normative modeling for novelty detection. Therefore, we aim to predict the masked fMRI 3D-volume from the presented visual stimuli in an encoding setting. To this end, we randomly selected 600 random pattern trials, for training the encoding model. The model then learns to represent this reference or normative class such that anomalous or abnormal samples can be detected and characterised. The rest of non-random patterns (720 trials) and random patterns (720 trials) are used for evaluating the encoding model and testing anomaly-detection performance, achieved by fitting a generalised extreme value distribution to the most deviating voxels. In our experiments, we use PCA to transform the fMRI data in the training set from the voxel space to $\mathbf{Z}$, and the resulting $P=10,25,50,100,250,500,1000$ PCA components are used as basis matrix $\mathbf{B}$ in the optimization and inference.

We benchmark the proposed method against the STGPR (\emph{i.e.}, mass-univariate) and MT-Kronprod models in terms of their runtime, performance of the regression, and quality of resulting normative models. In all models, we use a summation of a linear, a squared exponential, and a diagonal isotropic covariance functions for sample and task covariance matrices in order to accommodate both linear and non-linear relationships. In all cases, we use an isotropic Gaussian likelihood function. This likelihood function has different functionality in the STGPR versus MTGPR settings. In STGPR, it is defined independently for each voxel, thus it handles heteroscedastic, \emph{i.e.}, spatially varying noise. While in MTGPR a single noise parameter is shared for all voxels, hence it merely considers homoscedastic, \emph{i.e.}, spatially stationary, noise. The truncated Newton algorithm is used for optimizing the parameters. Table~\ref{tab:hyperparameters} summarizes the time complexity and the number of parameters of three benchmarked methods in our experiments.
\begin{table}[t]
\centering
\caption{Three benchmarked methods in our experiments.}
\vspace{-0.3cm}
\label{tab:hyperparameters}
\resizebox{0.995\textwidth}{!}{\begin{tabular}{|c|ccc|}
\hline
\textbf{Method} & \textbf{\begin{tabular}[c]{@{}c@{}}Time\\ Complexity\end{tabular}} & \textbf{\begin{tabular}[c]{@{}c@{}}No. \\ Parameters\end{tabular}} & \textbf{\begin{tabular}[c]{@{}c@{}}Parameter \\ Description\end{tabular}} \\ \hline
\textbf{STGPR} & $\mathcal{O}(N^3 T)$ & 21752 & \begin{tabular}[c]{@{}c@{}}1 for linear and 2 for squared exponential kernels, 1 for Gaussian likelihood; \\ multiplied by the number of tasks (5438).\end{tabular} \\ \hline
\textbf{MT-Kronprod} & $\mathcal{O}(T^3)$ & 9 & \begin{tabular}[c]{@{}c@{}}1 for linear, 2 for squared exponential, and 1 for diagonal isotropic kernels; multiplied \\ by 2 (for sample and task covariance functions);  plus 1 for Gaussian likelihood.\end{tabular} \\ \hline
\textbf{S-MTGPR} & $\mathcal{O}(N T^2)$ & 10 & \begin{tabular}[c]{@{}c@{}}Same as MT-Kronprod, plus 1 hyperparameter for the number of PCA bases.\end{tabular} \\ \hline
\end{tabular}}
\vspace{-0.75cm}
\end{table}

We use the coefficient of determination ($R^2$) to evaluate the explained variance by regression models. In normative modeling, the top 5\% values in normative probability maps are used to fit the generalized extreme value distribution (see~\cite{marquand2016understanding}). To evaluate resulting normative models, we employ area under the curve (AUC) to measure the performance of the model in distinguishing between normal (here random patterns) from abnormal samples (here non-random patterns). All the steps (random sampling, modeling, and evaluation) are repeated 10 times in order to estimate the mean and standard deviation of the runtime, $R^2$, and AUC. All experiments are performed on a system with Intel\textsuperscript \textregistered Xeon\textsuperscript \textregistered E5-1620 0 @3.60GHz CPU and 16GB of RAM.\footnote{\scriptsize The experimental codes are available at~\url{https://github.com/smkia/MTNorm}.\normalsize}
\vspace{-0.4cm}
\subsection{Results and Discussion} \label{subsec:results_discussion}
\begin{figure}[t!]
	\centering
	\includegraphics[width=0.995\textwidth]{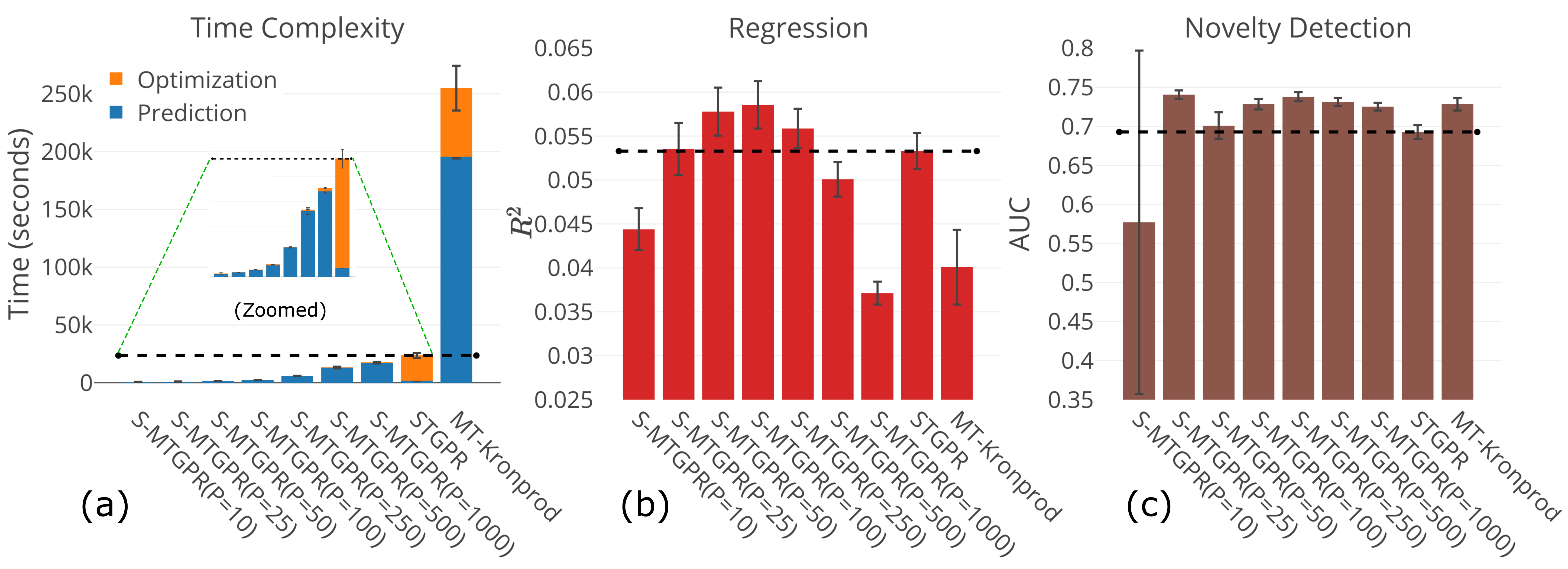}
	\vspace{-0.75cm}
	\caption{Comparison between S-MTGPR, STGPR, and MT-Kronprod in terms of: a) optimization and prediction runtime, b) average regression performance ($R^2$) across all voxels, and c) AUC in abnormal sample detection using normative modeling.}
	\label{fig:Comparison_bar_plots}
	\vspace{-0.6cm}
\end{figure}

Fig.~\ref{fig:Comparison_bar_plots} compares the runtime, $R^2$, and AUC of STGPR and MT-Kronprod, with those of S-MTGPR for different number of bases. As illustrated in Fig.~\ref{fig:Comparison_bar_plots}(a) S-MTGPR is faster than other approaches where the total runtime of MT-Kronprod (3 days) and STGPR (6 hours) can be  reduced to 16 minutes for $P=25$. This difference in runtime is even more pronounced in case of the optimization time where S-MTGPR is at least (for $P=1000$) 33 and 89 times faster than STGPR and MT-Kronprod, respectively. The multi-task approaches are slower than STGPR in the prediction phase mainly due to the mini-batch implementation of the prediction variance computation (to avoid memory overflow). Fig.~\ref{fig:Comparison_bar_plots}(b) shows this computational efficiency is achieved without penalty to the regression performance; where for certain number of bases the S-MTGPR shows equivalent and even better $R^2$ than STGPR and MT-Kronprod. Furthermore, Fig.~\ref{fig:Comparison_bar_plots}(c) demonstrates that multi-task learning, by considering spatial structures, generally provides a more accurate normative model of fMRI data in that it more accurately detects samples that were derived from a different distribution to those used to train the model. This fact is well-reflected in higher AUC values for S-MTGPR at $P=25,100,250,500,1000$. It is worthwhile to emphasize that these improvements are achieved by reducing the degree-of-freedom of the normative model from 21752 for STGPR to 10 for S-MTGPR (see Table~\ref{tab:hyperparameters}). 
\vspace{-0.35cm} 
\section{Conclusions and Future Work} \label{sec:conclusions}
\vspace{-0.1cm} 
Assuming a matrix normal distribution on a reduced latent output space, we introduced an efficient and scalable multi-task Gaussian process regression approach to learning complex association between external covariates and high-dimensional neuroimaging data. Our experiments on an fMRI dataset demonstrate the superiority of the proposed approach against other single-task and multi-task alternatives in terms of the computational time complexity. This superiority was achieved without compromising the regression performance, and even with higher sensitivity to abnormal samples in the normative modeling paradigm. Our methodological contribution advances the current practices in the normative modeling from the single-voxel modeling to multi-voxel structural learning. For future work, we will consider enriching the proposed approach by embedding more biologically meaningful basis functions~\cite{huertas2017bayesian}, structural modeling of non-stationary noise, and applying our method to clinical cohorts.
\vspace{-0.35cm} 
\bibliographystyle{splncs03}
\bibliography{references}
\newpage
\section*{Supplementary Materials}
\label{sec:supplementary}
Throughout the supplementary materials we use the same notation introduced in the main text.

\subsection*{Useful Equations}
\label{subsec:useful}
For $\mathbf{A} \in \mathbb{R}^{M \times N}$, $\mathbf{B} \in \mathbb{R}^{P \times Q}$, and $\mathbf{C}$, $\mathbf{D}$ (with appropriate size) we have:
\begin{enumerate}
\item $\mathbf{A = U_A S_A U_A^\top}$ is the eigenvalue decomposition of $\mathbf{A}$,
\item $\mathbf{(ACB)^{-1}=B^{-1}C^{-1}A^{-1}}$,
\item $\mathbf{(A \otimes B)(C \otimes D) = AC \otimes BD}$,
\item $\mathbf{(A \otimes B)^{-1} = A^{-1} \otimes B^{-1}}$,
\item the eigenvalue decomposition of $\mathbf{A \otimes B + I}$ is: \\ $\mathbf{(U_A \otimes U_B)(S_A \otimes S_B + I)(U_A^\top \otimes U_B^\top)}$,
\item $(\mathbf{A \otimes B}) vec(\mathbf{C}) = vec(\mathbf{BCA}^{\top})$,
\item $\ln \left | \mathbf{AC} \right | = \ln (\left | \mathbf{A} \right | \left | \mathbf{C} \right |) = \ln \left | \mathbf{A} \right | + \ln \left | \mathbf{C} \right |$,
\item for $\mathbf{C} \in \mathbb{R}^{N \times N}$, $\frac{\mathrm{d}}{\mathrm{d} x} \ln \left | \mathbf{C} \right | = Tr[\mathbf{C}^{-1} \frac{\mathrm{d} \mathbf{C}}{\mathrm{d} x}]$,
\item $Tr[\mathbf{ACBD}]=Tr[\mathbf{CBDA}]=Tr[\mathbf{BDAC}]=Tr[\mathbf{DACB}]$.
\end{enumerate}

\subsection*{Efficient Mean Prediction}
\label{subsec:mean_prediction}
Eq.~\ref{eq:efficient_prediction}(a) is derived from Eq.~\ref{eq:predictive_distribution}(a) as follows:
\small
\begin{eqnarray*} \label{eq:eff_mean_pred}
\begin{split}
vec(\mathbf{M^*}) & = (\mathbf{BCB^\top} \otimes \mathbf{R}^*)(\mathbf{BCB^\top} \otimes \mathbf{R} + \sigma^2 \mathbf{I})^{-1} vec(\mathbf{Y})  \\
& = (\mathbf{BCB}^\top \otimes \mathbf{R}^*) (\mathbf{B U_C S_C U_C^\top B^\top} \otimes \mathbf{U_R S_R U_R}^\top + \sigma^2 \mathbf{I})^{-1} vec(\mathbf{Y}) \\
& = (\mathbf{BCB}^\top \otimes \mathbf{R}^*) [(\mathbf{B U_C \otimes U_R}) (\mathbf{S_C \otimes S_R + \sigma^2 I})(\mathbf{U_C^\top B^\top \otimes U_R^\top})]^{-1} vec(\mathbf{Y}) \\
& = (\mathbf{BCB}^\top \otimes \mathbf{R}^*) (\mathbf{B U_C \otimes U_R}) (\mathbf{S_C \otimes S_R + \sigma^2 I})^{-1}(\mathbf{U_C^\top B^\top \otimes U_R^\top}) vec(\mathbf{Y}) \\
& = (\mathbf{BCB}^\top \otimes \mathbf{R}^*) (\mathbf{B U_C \otimes U_R}) (\mathbf{S_C \otimes S_R + \sigma^2 I})^{-1}vec(\mathbf{U_R^\top Y B U_C}) \\
& = (\mathbf{BC \underbrace{\mathbf{B^\top B}}_I U_C} \otimes \mathbf{R^* U_R}) \underbrace{diag[(\mathbf{S_C \otimes S_R + \sigma^2 I})^{-1}] \odot vec(\mathbf{U_R^\top Y B U_C})}_{vec({\tilde{\mathbf{Y}})}} \\ 
& = \mathbf{R^* U_R \tilde{\mathbf{Y}} \mathbf{U_C^\top C B^\top}} \quad .
\end{split}
\end{eqnarray*}
\normalsize

\subsection*{Efficient Variance Prediction}
\label{subsec:variance_prediction}
Eq.~\ref{eq:efficient_prediction}(b) is derived from Eq.~\ref{eq:predictive_distribution}(b) as follows:
\small
\begin{eqnarray*} \label{eq:eff_mean_pred}
\begin{split}
\mathbf{V}^* & = (\mathbf{BCB}^\top \otimes \mathbf{R}^{**})-(\mathbf{BCB}^\top \otimes \mathbf{R}^*)\underbrace{(\mathbf{BCB}^\top \otimes \mathbf{R} + \sigma^2 \mathbf{I})^{-1}}_{\mathbf{K}^{-1}}(\mathbf{BCB}^\top \otimes \mathbf{R}^{* \top}) \\
& = (\mathbf{BCB}^\top \otimes \mathbf{R}^{**})-(\mathbf{BCB}^\top \otimes \mathbf{R}^*)(\mathbf{B U_C \otimes U_R}) (\mathbf{S_C \otimes S_R + \sigma^2 I})^{-1} \\ & (\mathbf{U_C^\top B^\top \otimes U_R^\top}) (\mathbf{BCB}^\top \otimes \mathbf{R}^{* \top}) \\
& = (\mathbf{BCB}^\top \otimes \mathbf{R}^{**})-(\mathbf{BC U_C} \otimes \mathbf{R^* U_R})\underbrace{(\mathbf{S_C \otimes S_R + \sigma^2 I})^{-1}}_{\tilde{\mathbf{K}}^{-1} }(\mathbf{U_C^\top C B^\top \otimes U_R^\top} \mathbf{R}^{* \top}) \quad .
\end{split}
\end{eqnarray*}
\normalsize

\subsection*{Efficient Log Marginal Likelihood Evaluation}
\label{subsec:LML_evaluation}
Eq.~\ref{eq:LML} is derived as follows:
\small
\begin{eqnarray*} \label{eq:eff_LML}
\begin{split}
\mathcal{L} & = -\frac{N \times T}{2} \ln(2\pi)-\frac{1}{2}\ln\left | \mathbf{K} \right | - \frac{1}{2} vec(\mathbf{Y})^\top \mathbf{K}^{-1} vec(\mathbf{Y}) \\
& = -\frac{N \times T}{2} \ln(2\pi) - \frac{1}{2}\ln\left | \mathbf{BCB}^\top \otimes \mathbf{R} + \sigma^2 \mathbf{I}  \right | - \frac{1}{2} vec(\mathbf{Y})^\top (\mathbf{BCB}^\top \otimes \mathbf{R} + \sigma^2 \mathbf{I})^{-1} vec(\mathbf{Y}) \\
& = -\frac{N \times T}{2} \ln(2\pi)-\frac{1}{2}\ln\left | (\mathbf{B U_C \otimes U_R}) (\mathbf{S_C \otimes S_R + \sigma^2 I})(\mathbf{U_C^\top B^\top \otimes U_R^\top})  \right | \\ 
& - \frac{1}{2} vec(\mathbf{Y})^\top (\mathbf{B U_C \otimes U_R}) (\mathbf{S_C \otimes S_R + \sigma^2 I})^{-1}(\mathbf{U_C^\top B^\top \otimes U_R^\top})vec(\mathbf{Y}) \\
& = -\frac{N \times T}{2} \ln(2\pi)-\frac{1}{2} \underbrace{\ln\left | (\mathbf{U_C^\top B^\top \otimes U_R^\top})(\mathbf{B U_C \otimes U_R}) \right |}_{\ln \left | \mathbf{I}\right |=0} - \frac{1}{2} \ln\left |(\mathbf{S_C \otimes S_R + \sigma^2 I})\right |  \\ 
& - \frac{1}{2} vec(\mathbf{U_R^\top Y B U_C})^\top \underbrace{diag[(\mathbf{S_C \otimes S_R + \sigma^2 I})^{-1}] \odot vec(\mathbf{U_R^\top Y B U_C})}_{vec(\tilde{\mathbf{Y})}} \\
& = -\frac{N \times T}{2} \ln(2\pi) - \frac{1}{2} \ln \underbrace{\left |(\mathbf{S_C \otimes S_R + \sigma^2 I})\right |}_{\left | \tilde{\mathbf{K}} \right |} - \frac{1}{2} vec(\mathbf{U_R^\top Y B U_C})^\top vec(\tilde{\mathbf{Y}}) \quad .
\end{split}
\end{eqnarray*}
\normalsize

\subsection*{Derivatives of $\mathcal{L}$ with Respect to Parameters}
\label{subsec:gradients}
In the optimization process, the derivatives of $\mathcal{L}$ with respect to $\theta_{\mathbf{C}} \in \Theta_{\mathbf{C}}$, $\theta_{\mathbf{R}} \in \Theta_{\mathbf{R}}$, and $\theta_{\sigma^2} \in \Theta_{\sigma^2}$ can be efficiently computed as follows: 

\subsubsection*{Gradients of $\mathcal{L}$ with Respect to $\theta_\mathbf{C}$:}
\label{subsubsec:gradient_c}
\scriptsize
\begin{eqnarray*} \label{eq:derivatives_c}
\begin{aligned}
\frac{\partial \mathcal{L}}{\partial \theta_\mathbf{C}} = & -\frac{1}{2}diag(\tilde{\mathbf{K}}^{-1})^\top[diag(\mathbf{U_C^\top}\frac{\partial \mathbf{C}}{\partial \theta_\mathbf{C}}\mathbf{U_C}) \otimes diag(\mathbf{S_R})] 
 +\frac{1}{2} vec(\mathbf{\tilde Y})^\top vec(\mathbf{S_R \tilde Y U_C^\top \frac{\partial \mathbf{C}}{\partial \theta_\mathbf{C}}U_C}), 
\end{aligned}
\end{eqnarray*}
\normalsize
\noindent where the determinant term of the above equation is derived by computing the derivative of $\ln \left | \mathbf{K} \right |$:
\small
\begin{eqnarray*} \label{eq:determinant_c}
\begin{split}
\frac{\partial \ln \left | \mathbf{K} \right |}{\partial \theta_\mathbf{C}} & = \frac{\partial}{\partial \theta_\mathbf{C}} [\ln\left | \mathbf{BCB}^\top \otimes \mathbf{R} + \sigma^2 \mathbf{I}  \right |] = Tr[(\mathbf{BCB}^\top \otimes \mathbf{R} + \sigma^2 \mathbf{I})^{-1} \frac{\partial}{\partial \theta_\mathbf{C}}(\mathbf{BCB}^\top \otimes \mathbf{R} + \sigma^2 \mathbf{I})]  \\
& = Tr[(\mathbf{B U_C \otimes U_R}) (\mathbf{S_C \otimes S_R + \sigma^2 I})^{-1}(\mathbf{U_C^\top B^\top \otimes U_R^\top})(\mathbf{B \frac{\partial C}{\partial \theta_C} B^\top} \otimes \mathbf{R})] \\
& = Tr[(\mathbf{S_C \otimes S_R + \sigma^2 I})^{-1}(\mathbf{U_C^\top B^\top \otimes U_R^\top})(\mathbf{B \frac{\partial C}{\partial \theta_C} B^\top} \otimes \mathbf{R})(\mathbf{B U_C \otimes U_R})] \\
& = Tr[\mathbf{\tilde K}^{-1}(\mathbf{U_C^\top B^\top B \frac{\partial C}{\partial \theta_C} B^\top B U_C \otimes U_R^\top R U_R})] = Tr[\mathbf{\tilde K}^{-1}(\mathbf{U_C^\top \frac{\partial C}{\partial \theta_C} U_C \otimes S_R})] \\
& = diag(\mathbf{\tilde K}^{-1})^\top [diag(\mathbf{U_C^\top \frac{\partial C}{\partial \theta_C} U_C)} \otimes diag(\mathbf{S_R})] \quad ,
\end{split}
\end{eqnarray*}
\normalsize

\noindent and for the squared term we have:
\small
\begin{eqnarray*} \label{eq:squared_c}
\begin{split}
& \frac{\partial}{\partial \theta_\mathbf{C}} [vec(\mathbf{Y})^\top \mathbf{K}^{-1} vec(\mathbf{Y})] = \frac{\partial}{\partial \theta_\mathbf{C}} [vec(\mathbf{Y})^\top (\mathbf{BCB}^\top \otimes \mathbf{R} + \sigma^2 \mathbf{I})^{-1} vec(\mathbf{Y})] \\
& = - vec(\mathbf{Y})^\top (\mathbf{BCB}^\top \otimes \mathbf{R} + \sigma^2 \mathbf{I})^{-1} [\frac{\partial}{\partial \theta_\mathbf{C}}(\mathbf{BCB}^\top \otimes \mathbf{R} + \sigma^2 \mathbf{I})] (\mathbf{BCB}^\top \otimes \mathbf{R} + \sigma^2 \mathbf{I})^{-1} vec(\mathbf{Y}) \\
& = - vec(\mathbf{Y})^\top (\mathbf{B U_C \otimes U_R}) (\mathbf{S_C \otimes S_R + \sigma^2 I})^{-1}(\mathbf{U_C^\top B^\top \otimes U_R^\top}) (\mathbf{B\frac{\partial C }{\partial \theta_C} B^\top} \otimes \mathbf{R}) \\
& (\mathbf{B U_C \otimes U_R}) (\mathbf{S_C \otimes S_R + \sigma^2 I})^{-1}(\mathbf{U_C^\top B^\top \otimes U_R^\top}) vec(\mathbf{Y}) \\
& = - [vec(\mathbf{U_R^\top Y B U_C})^\top \odot diag(\tilde{\mathbf{K}}^{-1})] (\mathbf{U_C^\top B^\top B \frac{\partial C}{\partial \theta_C} B^\top B U_C \otimes U_R^\top R U_R}) \\ & [\underbrace{diag(\tilde{\mathbf{K}}^{-1}) \odot vec(\mathbf{U_R^\top Y B U_C})}_{vec(\mathbf{\tilde Y})}]
 = - vec(\mathbf{\tilde Y})^\top vec(\mathbf{S_R \tilde{Y} U_C^\top \frac{\partial C}{\partial \theta_C} U_C}) \quad .
\end{split}
\end{eqnarray*}
\normalsize

\subsubsection*{Gradients of $\mathcal{L}$ with Respect to $\theta_\mathbf{R}$:}
\label{subsubsec:gradient_c}
\scriptsize
\begin{eqnarray} \label{eq:derivatives_r}
\begin{aligned}
\frac{\partial \mathcal{L}}{\partial \theta_\mathbf{R}} = & -\frac{1}{2}diag(\tilde{\mathbf{K}}^{-1})^\top[diag(\mathbf{S_C}) \otimes diag(\mathbf{U_R^\top}\frac{\partial \mathbf{R}}{\partial \theta_\mathbf{R}}\mathbf{U_R})]
 +\frac{1}{2} vec(\mathbf{\tilde Y})^\top vec(\mathbf{U_R^\top \frac{\partial \mathbf{R}}{\partial \theta_\mathbf{R}}U_R \tilde Y S_C}),
\end{aligned}
\end{eqnarray}
\normalsize
\noindent where the determinant term of the above equation is derived by computing the derivative of $\ln \left | \mathbf{K} \right |$:
\small
\begin{eqnarray*} \label{eq:determinant_r}
\begin{split}
\frac{\partial \ln \left | \mathbf{K} \right |}{\partial \theta_\mathbf{R}} & = \frac{\partial}{\partial \theta_\mathbf{R}} [\ln\left | \mathbf{BCB}^\top \otimes \mathbf{R} + \sigma^2 \mathbf{I} \right |] = Tr[(\mathbf{BCB}^\top \otimes \mathbf{R} + \sigma^2 \mathbf{I})^{-1} \frac{\partial}{\partial \theta_\mathbf{R}}(\mathbf{BCB}^\top \otimes \mathbf{R} + \sigma^2 \mathbf{I})]  \\
& = Tr[(\mathbf{B U_C \otimes U_R}) (\mathbf{S_C \otimes S_R + \sigma^2 I})^{-1}(\mathbf{U_C^\top B^\top \otimes U_R^\top})(\mathbf{B C B^\top} \otimes \mathbf{\frac{\partial R}{\partial \theta_R}})] \\
& = Tr[(\mathbf{S_C \otimes S_R + \sigma^2 I})^{-1}(\mathbf{U_C^\top B^\top \otimes U_R^\top})(\mathbf{B C B^\top} \otimes \mathbf{\frac{\partial R}{\partial \theta_R}})(\mathbf{B U_C \otimes U_R})] \\
& = Tr[\mathbf{\tilde K}^{-1}(\mathbf{U_C^\top B^\top B C B^\top B U_C \otimes U_R^\top \frac{\partial R}{\partial \theta_R} U_R})] = Tr[\mathbf{\tilde K}^{-1}(\mathbf{S_C \otimes  U_R^\top \frac{\partial R}{\partial \theta_R} U_R})] \\
& = diag(\mathbf{\tilde K}^{-1})^\top [diag(\mathbf{S_C}) \otimes diag(\mathbf{ U_R^\top \frac{\partial R}{\partial \theta_R} U_R})] \quad ,
\end{split}
\end{eqnarray*}
\normalsize

\noindent and for the squared term we have:
\small
\begin{eqnarray*} \label{eq:squared_r}
\begin{split}
& \frac{\partial}{\partial \theta_\mathbf{R}} [vec(\mathbf{Y})^\top \mathbf{K}^{-1} vec(\mathbf{Y})] = \frac{\partial}{\partial \theta_\mathbf{R}} [vec(\mathbf{Y})^\top (\mathbf{BCB}^\top \otimes \mathbf{R} + \sigma^2 \mathbf{I})^{-1} vec(\mathbf{Y})] \\
& = - vec(\mathbf{Y})^\top (\mathbf{BCB}^\top \otimes \mathbf{R} + \sigma^2 \mathbf{I})^{-1} [\frac{\partial}{\partial \theta_\mathbf{R}}(\mathbf{BCB}^\top \otimes \mathbf{R} + \sigma^2 \mathbf{I})] (\mathbf{BCB}^\top \otimes \mathbf{R} + \sigma^2 \mathbf{I})^{-1} vec(\mathbf{Y}) \\
& = - vec(\mathbf{Y})^\top (\mathbf{B U_C \otimes U_R}) (\mathbf{S_C \otimes S_R + \sigma^2 I})^{-1}(\mathbf{U_C^\top B^\top \otimes U_R^\top}) (\mathbf{BCB^\top \otimes \frac{\partial R}{\partial \theta_R}}) \\
& (\mathbf{B U_C \otimes U_R}) (\mathbf{S_C \otimes S_R + \sigma^2 I})^{-1}(\mathbf{U_C^\top B^\top \otimes U_R^\top}) vec(\mathbf{Y}) \\
& = - [vec(\mathbf{U_R^\top Y B U_C})^\top \odot diag(\tilde{\mathbf{K}}^{-1})] (\mathbf{U_C^\top B^\top BCB^\top B U_C \otimes U_R^\top \frac{\partial R}{\partial \theta_R} U_R}) \\ 
& [\underbrace{diag(\tilde{\mathbf{K}}^{-1}) \odot vec(\mathbf{U_R^\top Y B U_C})}_{vec(\mathbf{\tilde Y})}]
 = - vec(\mathbf{\tilde Y})^\top vec(\mathbf{U_R^\top \frac{\partial R}{\partial \theta_R} U_R \tilde{Y} S_C}) \quad .
\end{split}
\end{eqnarray*}
\normalsize

\subsubsection*{Gradients of $\mathcal{L}$ with Respect to $\theta_{\sigma^2}$:}
\label{subsubsec:gradient_s}
\small
\begin{eqnarray} \label{eq:derivatives_s}
\begin{aligned}
\frac{\partial \mathcal{L}}{\partial \theta_{\sigma^2}} = & -\frac{1}{2} \frac{\partial \sigma^2}{\partial \theta_{\sigma^2}} [Tr[\tilde{\mathbf{K}}^{-1}] + vec(\mathbf{\tilde Y})^\top vec(\mathbf{\tilde Y})] \quad ,
\end{aligned}
\end{eqnarray}
\normalsize
\noindent where the determinant term of the above equation is derived by computing the derivative of $\ln \left | \mathbf{K} \right |$:
\small
\begin{eqnarray*} \label{eq:determinant_s}
\begin{split}
\frac{\partial \ln \left | \mathbf{K} \right |}{\partial \theta_{\sigma^2}} & = \frac{\partial}{\partial \theta_{\sigma^2}} [\ln\left | \mathbf{BCB}^\top \otimes \mathbf{R} + \sigma^2 \mathbf{I} \right |] = Tr[(\mathbf{BCB}^\top \otimes \mathbf{R} + \sigma^2 \mathbf{I})^{-1} \frac{\partial \sigma^2}{\partial \theta_{\sigma^2}}]  \\
& = \frac{\partial \sigma^2}{\partial \theta{\sigma^2}}Tr[(\mathbf{B U_C \otimes U_R}) (\mathbf{S_C \otimes S_R + \sigma^2 I})^{-1}(\mathbf{U_C^\top B^\top \otimes U_R^\top})] \\ 
& = \frac{\partial \sigma^2}{\partial \theta{\sigma^2}} Tr[(\mathbf{S_C \otimes S_R + \sigma^2 I})^{-1}(\mathbf{U_C^\top B^\top \otimes U_R^\top})(\mathbf{B U_C \otimes U_R})] \\
& = \frac{\partial \sigma^2}{\partial \theta{\sigma^2}} Tr[\mathbf{\tilde K}^{-1}(\mathbf{U_C^\top B^\top B U_C \otimes U_R^\top  U_R})] = \frac{\partial \sigma^2}{\partial \theta{\sigma^2}} Tr[\mathbf{\tilde K}^{-1}] \quad ,
\end{split}
\end{eqnarray*}
\normalsize

\noindent and for the squared term we have:
\small
\begin{eqnarray*} \label{eq:squared_s}
\begin{split}
& \frac{\partial}{\partial \theta_{\sigma^2}} [vec(\mathbf{Y})^\top \mathbf{K}^{-1} vec(\mathbf{Y})] = \frac{\partial}{\partial \theta_{\sigma^2}} [vec(\mathbf{Y})^\top (\mathbf{BCB}^\top \otimes \mathbf{R} + \sigma^2 \mathbf{I})^{-1} vec(\mathbf{Y})] \\
& = - vec(\mathbf{Y})^\top (\mathbf{BCB}^\top \otimes \mathbf{R} + \sigma^2 \mathbf{I})^{-1} [\frac{\partial}{\partial \theta_{\sigma^2}}(\mathbf{BCB}^\top \otimes \mathbf{R} + \sigma^2 \mathbf{I})] (\mathbf{BCB}^\top \otimes \mathbf{R} + \sigma^2 \mathbf{I})^{-1} vec(\mathbf{Y}) \\
& = - vec(\mathbf{Y})^\top (\mathbf{B U_C \otimes U_R}) (\mathbf{S_C \otimes S_R + \sigma^2 I})^{-1}(\mathbf{U_C^\top B^\top \otimes U_R^\top}) \frac{\partial \sigma^2}{\partial \theta_{\sigma^2}} \\
& (\mathbf{B U_C \otimes U_R}) (\mathbf{S_C \otimes S_R + \sigma^2 I})^{-1}(\mathbf{U_C^\top B^\top \otimes U_R^\top}) vec(\mathbf{Y}) \\
& = - \frac{\partial \sigma^2}{\partial \theta_{\sigma^2}} [vec(\mathbf{U_R^\top Y B U_C})^\top \odot diag(\tilde{\mathbf{K}}^{-1})] (\mathbf{U_C^\top B^\top B U_C \otimes U_R^\top U_R}) \\
& [\underbrace{diag(\tilde{\mathbf{K}}^{-1}) \odot vec(\mathbf{U_R^\top Y B U_C})}_{vec(\mathbf{\tilde Y})}]
= - \frac{\partial \sigma^2}{\partial \theta_{\sigma^2}} vec(\mathbf{\tilde Y})^\top vec(\mathbf{\tilde Y}) \quad .
\end{split}
\end{eqnarray*}
\normalsize

\subsection*{Normative Modeling}
\label{subsec:normative_modeling}
Let $\hat{y}_{ij}$ and $\sigma^2_{ij}$ be the prediction mean and variance of the $i$th test sample at the $j$th voxel. Further, let $\sigma^2_{nj}$ be the variance of the noise that is estimated by GPR at the $j$th voxel. Then the normative probability map (NPM) for the $i$th sample at $j$th voxel is defined as follows:
\small
\begin{eqnarray*} \label{eq:NPM}
NPM_{ij}=\frac{y_{ij}-\hat{y}_{ij}}{\sqrt{\sigma^2_{ij}+\sigma^2_{nj}}}
\quad ,
\end{eqnarray*}
\normalsize
\noindent where $y_{ij}$ is the true output. Having computed NPMs for all samples and brain locations, the abnormality index of each sample can be computed by fitting a generalized extreme value distribution (GEVD). We fit GEVD on the distribution of robust means of top 5\% voxels (in absolute value) across all NPMs. The resulting distribution is used to compute the probability of each sample being abnormal.

\end{document}